%% file: iclr2024_conference.tex
\documentclass{article} 
\usepackage{iclr2024_conference,times}

\input{math_commands.tex}

\usepackage{hyperref}
\usepackage{url}
\usepackage{cleveref}
\usepackage{graphicx}
\usepackage{array}
\usepackage{subcaption} 
\usepackage{caption}
\usepackage{comment}
\usepackage{xcolor}

\usepackage{tcolorbox}
\usepackage{listings}

\usepackage{xcolor}

\definecolor{citecolor}{RGB}{56,75,178}
\definecolor{linkcolor}{RGB}{105,52,96}

\hypersetup{
    colorlinks,
    linkcolor=linkcolor,
    citecolor=citecolor,
    urlcolor={blue!80!black}
}

\definecolor{codebg}{rgb}{0.95,0.95,0.95}
\definecolor{codeborder}{rgb}{0.8,0.8,0.8}

\crefname{lstlisting}{listing}{listings}
\Crefname{lstlisting}{Listing}{Listings}

\newcommand{\benchmark}{\textsc{RES-Q}}

\Crefname{section}{\S}{\S\S}
\crefformat{section}{\S#2#1#3}
\Crefname{figure}{Figure}{Figures}
\crefformat{figure}{Figure #2#1#3}
\Crefname{Figure}{Figure}{Figures}
\Crefname{Table}{Table}{Tables}
\crefformat{table}{Table #2#1#3}

\title{RES-Q: Evaluating Code-Editing Large Language Model Systems at the Repository Scale}

\author{%
    Beck LaBash$^{2}$\thanks{Work completed while at Qurrent AI} \quad August Rosedale$^{1}$\thanks{Correspondence to \texttt{\href{mailto:august@qurrent.ai}{august@qurrent.ai}}} \quad Alex Reents$^{1}$ \quad Lucas Negritto$^{1}$ \quad Colin Wiel$^{1}$ \\[0.2cm]
    $^1$Qurrent AI \quad $^2$Northeastern University \\
}

\iclrfinalcopy 
\begin{document}

\maketitle
\begin{abstract}
The instruction-following ability of Large Language Models (LLMs) has cultivated a class of LLM-based systems capable of approaching complex tasks such as making edits to large code repositories. Due to the high sensitivity and unpredictability of LLM behavior in response to changes in prompting, robust evaluation tools are needed to drive future iteration of these systems. We propose \benchmark{}, a natural language instruction-based benchmark for evaluating \textbf{\underline{R}}epository \textbf{\underline{E}}diting \textbf{\underline{S}}ystems, which consists of 100 handcrafted repository editing tasks derived from real GitHub commits. Given an edit instruction and a code repository, RES-Q evaluates an LLM system's ability to interpret the instruction, navigate the repository to gather relevant information, and construct an appropriate edit that satisfies the specified criteria. We argue that evaluating LLMs in this way addresses issues with traditional benchmarks and provides a more holistic assessment of a model's abilities. We evaluate various state-of-the-art LLMs as language agents in a repository-editing system built on Qurrent OS, our language agent development software.
Despite their 1\% pass@1 performance difference on HumanEval, we find Claude Sonnet 3.5 outperforms GPT-4o by 12\% pass@1 on RES-Q, indicating RES-Q's capacity to differentiate model capability as traditional benchmarks approach saturation.
We further investigate token efficiency, performance relationships with existing benchmarks, and interesting disparities between closed and open-source LLMs.
Code and dataset are available at \href{https://github.com/Qurrent-AI/RES-Q}{this https URL}.
\end{abstract}

\section{Introduction}

\subsection{Automating Software Development with LLMs}
The recent increase in the versatile capability of Large Language Models (LLMs) has given rise to a class of systems referred to as LLM-based agents, or language agents for short \citep{xi2023rise}. In general, a language agent is LLM augmented by planning, memory, and tool-use capabilities \citep{weng2023agent}. Under this umbrella, there exists a set of language agent systems that seek to automate software development tasks. \citet{gpt-engineer} employs a user feedback loop to clarify task details. \citet{AutoGPT} decomposes its task into a tree of subtasks. \citet{hong2023metagpt} coordinates multiple language agents working as different roles in a software assembly line. Each with a unique strategy, the ability of these systems to develop simple web apps has been an impressive enough spectacle to raise questions about the future role of humans in software development. 

However, the majority of software development work involves extending, editing, and maintaining existing large codebases rather than creating new ones from scratch. In these more realistic settings, the challenge of managing and reasoning across large contexts has been prohibitive to similarly sensational performance. Nonetheless, recent efforts such as CodeR \citep{chen2024coder} and Aider \citep{aider} have shown the potential for near-term improvement in these areas with LLM-based systems. As models and systems become more capable, it is crucial to have robust evaluation frameworks that can accurately assess their performance on real-world software development tasks.

\subsection{Evaluation of LLMs}

Historically, the performance of an LLM has been characterized by its accuracy on a set of annotated NLP benchmarks such as BIG-Bench Hard \citep{suzgun2022challenging} and MMLU \citep{hendrycks2021measuring} which test knowledge and problem-solving capabilities. However, these benchmarks are primarily designed to evaluate LLMs in zero-shot or few-shot settings \citep{hendrycks2021measuring}, measuring only the knowledge acquired by the model during pre-training. Additionally, \citet{Ott2022MappingGD} shows that the majority of NLP benchmarks quickly tend towards saturation, and \citet{sainz-etal-2023-nlp} demonstrates that data contamination can compromise traditional annotated benchmarks, leading to an overestimation of model performance. 

Evaluating LLMs based on their performance as a language agent addresses the identified shortcomings of current annotated NLP benchmarks. Specifically, the utility of a particular LLM as a language agent depends on its ability to precisely follow instructions, understand roles, and make informed, sequential decisions, thereby providing a more holistic evaluation of the model. Additionally, complex tasks undertaken by language agents inherently present higher difficulty levels, reducing the risk of benchmark saturation observed in traditional evaluations. Furthermore, the novelty and diversity of language agent trajectories significantly mitigate the risk of data contamination, as each trajectory is influenced by unique interactions and evolving contexts. 

\subsection{Contributions}

In this paper, we introduce \benchmark{}, a hand-crafted dataset consisting of 100 natural language code repository edit instructions paired with corresponding test suites. \benchmark{}'s instructions are derived from actual commits made to GitHub repositories, ensuring their relevance and realism. Concretely, \benchmark{} serves as a robust evaluation tool for assessing the performance of LLM-based systems on real-world software development tasks, particularly those involving the extension and modification of large code repositories. More generally, \benchmark{} provides an opportunity to evaluate the objective, binary success of an LLM working as a language agent on a complex task, thereby offering a more comprehensive indication of its capabilities compared to traditional benchmarks. To this end, we conduct an assessment of state-of-the-art models within a code-repository editing system built on Qurrent OS, our proprietary language agent development software, and compare the results with traditional benchmarks. To complement our dataset, we also release a submission environment to evaluate the correctness of repository changes for a given task, promoting broader community engagement.

To summarize, we make the following contributions:

\begin{enumerate}
    \item \textbf{Introduction of \benchmark{} Dataset:} We introduce \benchmark{}, an instruction-based dataset of codebase edits derived from actual GitHub commits, designed to evaluate the performance of LLM-based systems on real-world software development tasks (\Cref{sec:dataset}).
    
    \item \textbf{Evaluation of SOTA LLMs:} We evaluate state-of-the-art LLMs within a codebase-editing system built on Qurrent OS, our language agent development software, and investigate correlations with performance on traditional benchmarks (\Cref{sec:evaluation}).
    
    \item \textbf{Submission Environment:} We release a submission environment with a simple interface for evaluating the correctness of codebase changes for a given \benchmark{} task, facilitating the evaluation of other systems and encouraging broader community engagement (\Cref{subsec:submission-env}).
\end{enumerate}

\section{Related Work}

\textbf{Code Editing Benchmarks.} SWE-Bench \citep{jimenez2024swebench} is a Python codebase editing benchmark constructed automatically from GitHub Issues and their accompanying Pull Requests. Designed to closely emulate real-world software development tasks, SWE-Bench uses raw GitHub Issue threads as task instructions. In contrast, \benchmark{}'s instructions consist of intentionally vague, mostly single-sentence imperative instructions, similar to the \textit{lazy} edit instructions described in \citet{cassano2024edit}. While GitHub Issues are indeed a realistic artifact of software development, engineers initially approaching a task will not often have the prescriptiveness provided by GitHub Issues. \benchmark{}'s instructions capture this reality by testing the ability to reason with ambiguity present in task definitions. Additionally, by avoiding the direct use of GitHub data in task instructions, \benchmark{} reduces the risk of data contamination for models trained on GitHub. SWE-Bench filters for task instances with \textit{fail-to-pass} tests, unit tests introduced in the PR that fail on the repository before the PR and pass after \citep{jimenez2024swebench}. While this ensures that changes meet a minimum threshold of correctness, it does not guarantee comprehensive verification.  Unit tests added during PRs often cover specific edge cases or a limited subset of the new functionality, potentially leaving significant portions of the code unverified. In contrast, \benchmark{} includes a meticulously handcrafted test suite for each task to ensure exhaustive verification of solutions. Moreover, while SWE-Bench focuses solely on Python codebases, \benchmark{} includes tasks for both Python and JavaScript codebases, broadening the scope of evaluation.

CanItEdit \citep{cassano2024edit} is a handcrafted dataset of 105 instructional code editing problems, each centered on single Python files. CanItEdit's \textit{lazy} instructions model is characterized by its intentional brevity and vagueness, an approach that has been adopted by \benchmark{}. Both benchmarks use handcrafted test suites to validate task solutions. CanItEdit explicitly provides the file to be edited, isolating the precise code editing task within a single file context. In contrast, \benchmark{} provides the entire codebase as context, requiring deliberation about which files need to be edited, and thereby presenting a broader software development task for evaluation. In terms of task complexity, CanItEdit tasks involve changing only one file, while \benchmark{} tasks often require edits across multiple files. Additionally, the edits in \benchmark{} are generally more extensive and complex compared to those in CanItEdit, as indicated by the higher median Levenshtein distance (see Table~\ref{table:dataset-stats}). This highlights \benchmark{}'s focus on broader context tasks, aligning more closely with real-world software development challenges.

Overall, \benchmark{} integrates the commit-based, handcrafted nature of CanItEdit with the real-world task environment of SWE-bench. This combination aims to provide a comprehensive and realistic evaluation framework for LLM-based systems, facilitating their assessment on practical software development tasks and the broader capabilities of the system's LLM.

\section{RES-Q}\label{sec:dataset}

\begin{table}[b]
\begin{minipage}{.49\linewidth}
    \begin{minipage}{\textwidth}
        \centering
        \includegraphics[width=\textwidth]{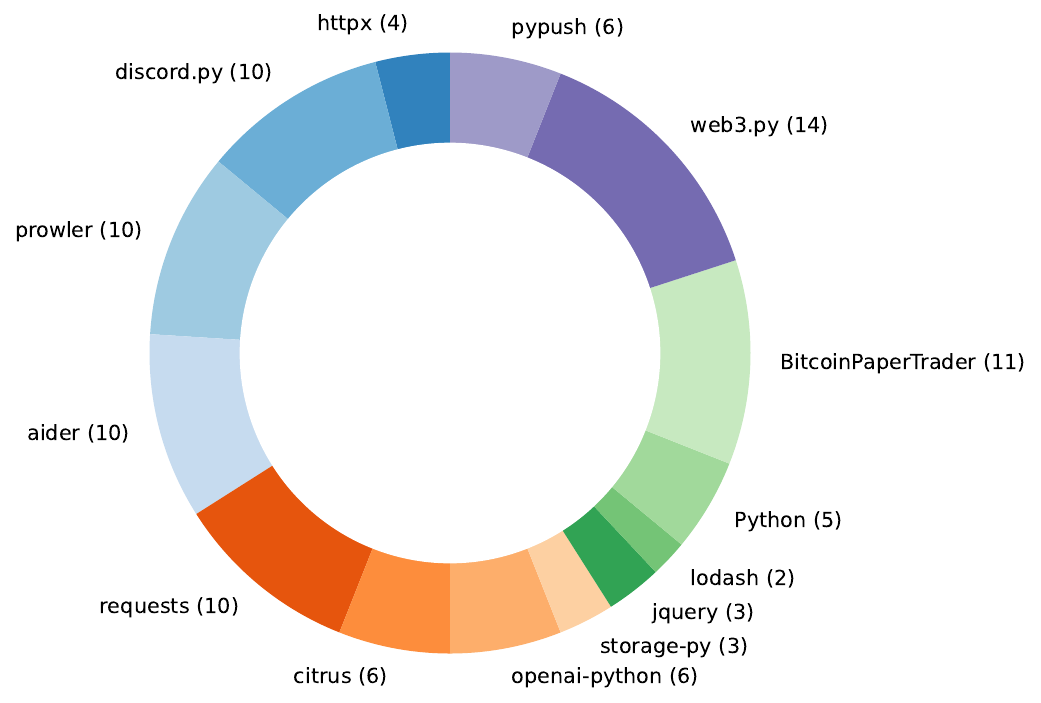}
        \captionof{figure}{Distribution of \benchmark{} tasks across different repositories.}
        \label{fig:task-repos}
    \end{minipage}
\end{minipage}
\hfill
\begin{minipage}{.48\linewidth}
    \begin{minipage}{\textwidth}
        \centering
        \caption{Dataset statistics for \benchmark{}.}
        \renewcommand{\arraystretch}{1.2}
        \resizebox{\textwidth}{!}{
        \small
        \begin{tabular}{|l|c|}
        \hline
        \multicolumn{2}{|c|}{\textbf{\benchmark{} Dataset Statistics}} \\
        \hline
        Total Tasks & 100 \\
        \hline

        \textbf{Repository} & \textbf{Q1 $\mid$ Median $\mid$ Q3} \\
        \hline
        Lines of Code (LoC) & 4170 $\mid$ 6468 $\mid$ 42793 \\
        \cline{2-2}
        Files & 20 $\mid$ 55 $\mid$ 395 \\
        \hline

        \textbf{Ground Truth Patch} & \textbf{Q1 $\mid$ Median $\mid$ Q3} \\
        \hline
        Levenshtein Distance & 119.5 $\mid$ 730 $\mid$ 1720 \\
        \hline
        Modified LoC & 6 $\mid$ 27.5 $\mid$ 63.7 \\
        \hline
        Modified Files & 1 $\mid$ 1 $\mid$ 3 \\
        \hline
        Tokens & 224.5 $\mid$ 523 $\mid$ 885.25\\
        \hline

        \textbf{Instruction} & \textbf{Mean $\pm$ Std. Dev.} \\
        \hline
        Tokens & 70.45 $\pm$ 36.68 \\
        \hline
        \end{tabular}
        }
        \label{table:dataset-stats}
    \end{minipage}
\end{minipage}
\vspace{-0.5em}
\end{table}

\benchmark{} is a codebase editing benchmark based on compact, natural language instructions. Given an edit instruction and a codebase, the task is to make an edit to the codebase that satisfies the criteria set by the instruction. \Cref{table:dataset-stats} presents general dataset statistics for \benchmark{}.

\subsection{Benchmark Construction}
\begin{figure}[t]
\centering
\begin{minipage}[t]{0.5\textwidth}
\centering
\includegraphics[width=0.9\linewidth]{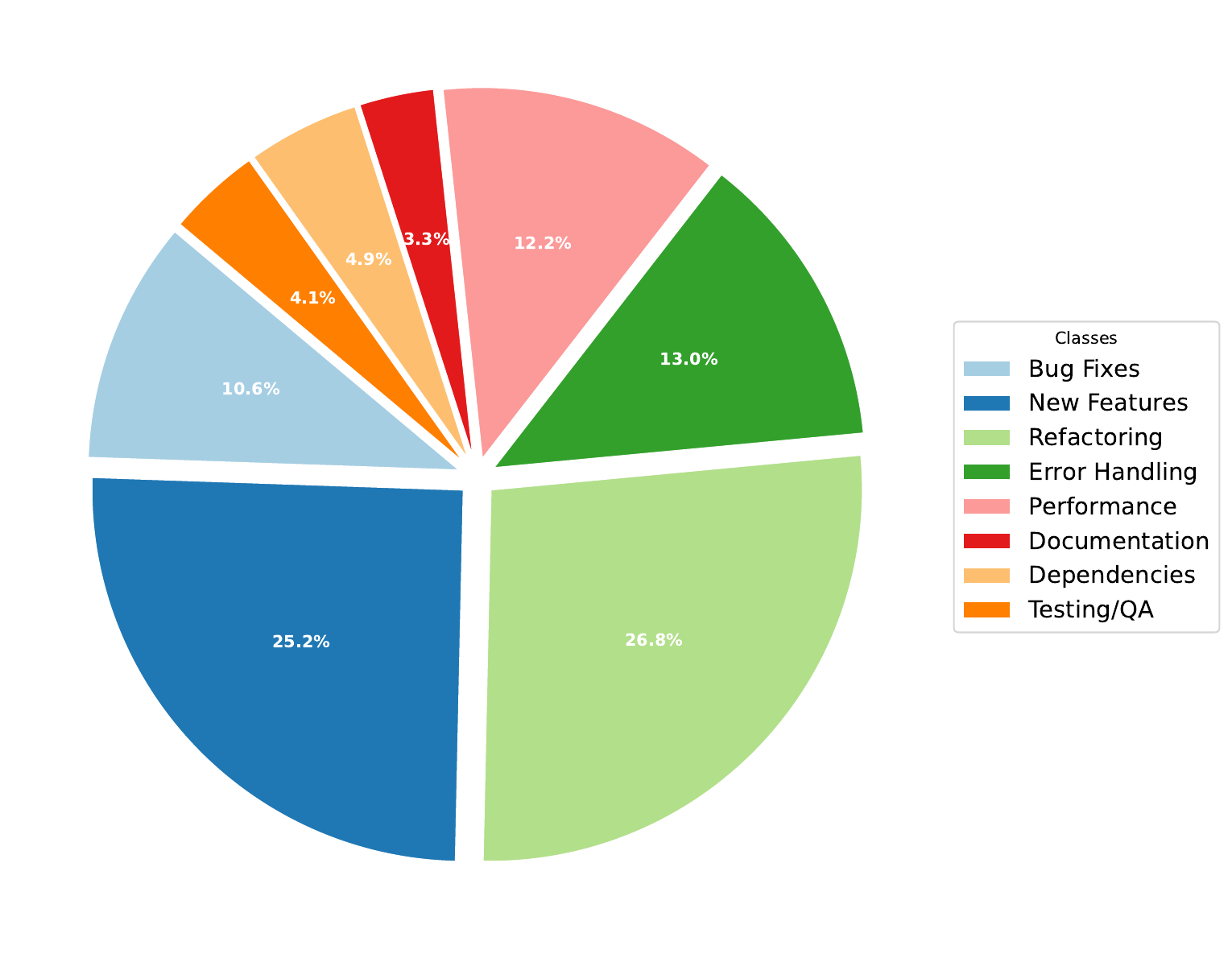}
\captionsetup{width=0.9\linewidth}
\caption{Distribution of \benchmark{} tasks across different instruction types.}
\label{fig:objective-class}
\end{minipage}%
\begin{minipage}[t]{0.5\textwidth}
\centering
\includegraphics[width=0.9\linewidth]{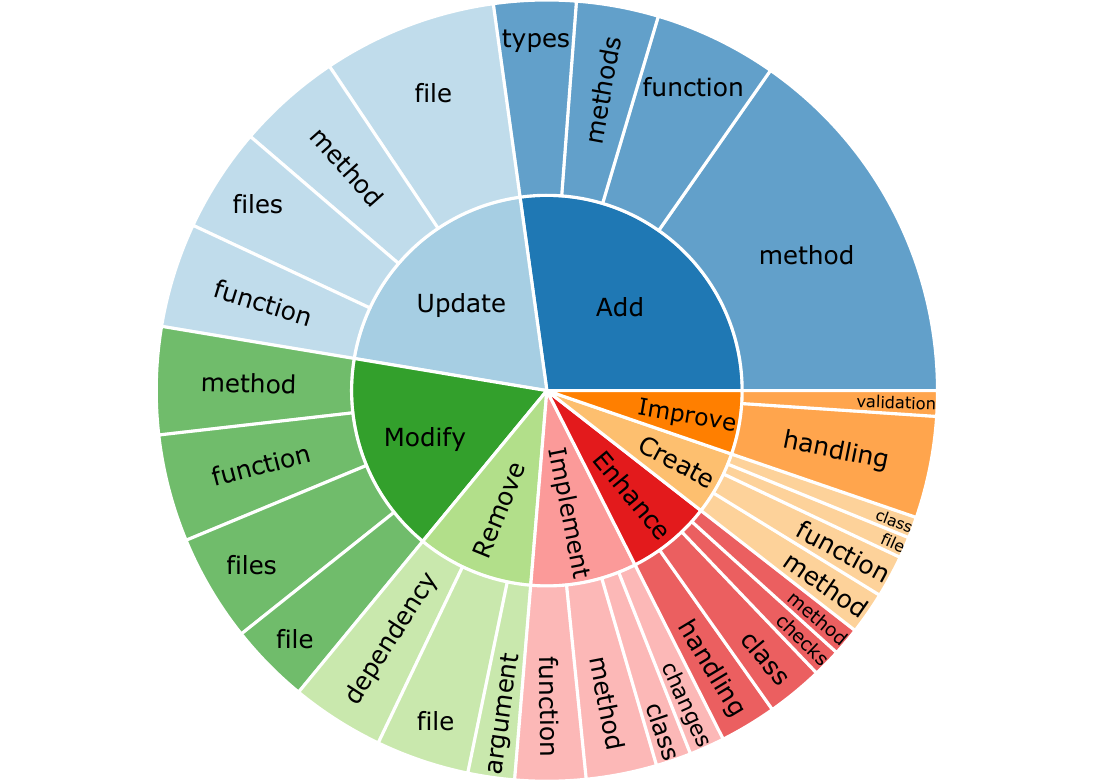}
\captionsetup{width=0.9\linewidth}
\caption{The 8 most frequent primary action verbs with their corresponding 4 most frequent target nouns occurring in \benchmark{} task instructions.}
\label{fig:objective-verbs}
\end{minipage}
\end{figure}

To construct \benchmark{}, we selected 14 GitHub repositories of varying size and popularity. We combed through each repository's commit history and selected a representative collection of changes whose distribution mirrors real-world software development practices. The distribution of \benchmark{} tasks across these repositories is shown in \Cref{fig:task-repos}, and \Cref{fig:objective-class} illustrates the distribution of tasks across different classifications. For each selected change, we recorded the \textit{base commit} (the starting point of the task), as well as its associated \textit{solution commit}, (the ground truth change). It's important to note that the \textit{solution commit} is not necessarily the direct child of the \textit{base commit}; intermediate commits may exist between them.  

We crafted an edit instruction and a Python-based testing environment for each selected change. Each testing environment includes a Python version, a requirements file, and a test suite. As an additional verification step, we used our submission environment to ensure that the test suite fails on the \textit{base commit}, and passes on a patch generated from the diff between the \textit{base commit} and \textit{solution commit}.

We created forks of each repository included in the benchmark to ensure consistency over time.

\subsection{Task Formulation}

\textbf{Prompt.} The prompt for each \benchmark{} task includes: 

\begin{enumerate}
    \item A natural language instruction describing the desired change.
    \item The entire codebase checked out at the task's \textit{base commit}.
\end{enumerate}

In practice, the \benchmark{} dataset provides the forked GitHub repository URL and \textit{base commit} hash as a reference to the codebase.

\textbf{Submission.} A \benchmark{} task submission consists of an edit to the codebase which makes the change described in the instruction. The submission environment, described in \Cref{subsec:submission-env}, represents this edit as a unified diff patch file.

\textbf{Evaluation Metric.} Tasks are evaluated based on binary success. A task is considered successful if the provided patch applies successfully to the repository, and the resulting repository passes the task's test suite. 
The primary metric for \benchmark{} is \textit{pass@k} which measures the likelihood of producing at least one successful completion per task, given a fixed number \textit{k} attempts. In the case where each task is attempted once, \textit{pass@1} represents the proportion of successful tasks to total tasks.


An example \benchmark{} task is provided in \Cref{appx:example-task}.

\subsection{Submission Environment}\label{subsec:submission-env}

We implement a submission environment to facilitate the easy evaluation of solutions to \benchmark{} tasks generated by other LLM systems. The environment accepts a \textit{submission}, consisting of a task identifier and unified diff patch file, and produces a boolean indicating the success of the submission, a feedback message (one of \textit{PASS}, \textit{FAIL}, \textit{PATCH FAILED}, \textit{TIMED OUT}), and feedback from the test suite. The environment is implemented asynchronously, enabling concurrent evaluation of multiple submissions while maintaining complete isolation of each one. This design ensures both scalability and reliability in the evaluation process. See \Cref{appx:submission-env} for a diagram describing the high-level architecture of the submission environment.

\section{Evaluation}\label{sec:evaluation}

\begin{figure}[t]
    \centering
    \begin{subfigure}[t]{0.48\textwidth}
        \centering
        \includegraphics[width=\linewidth]{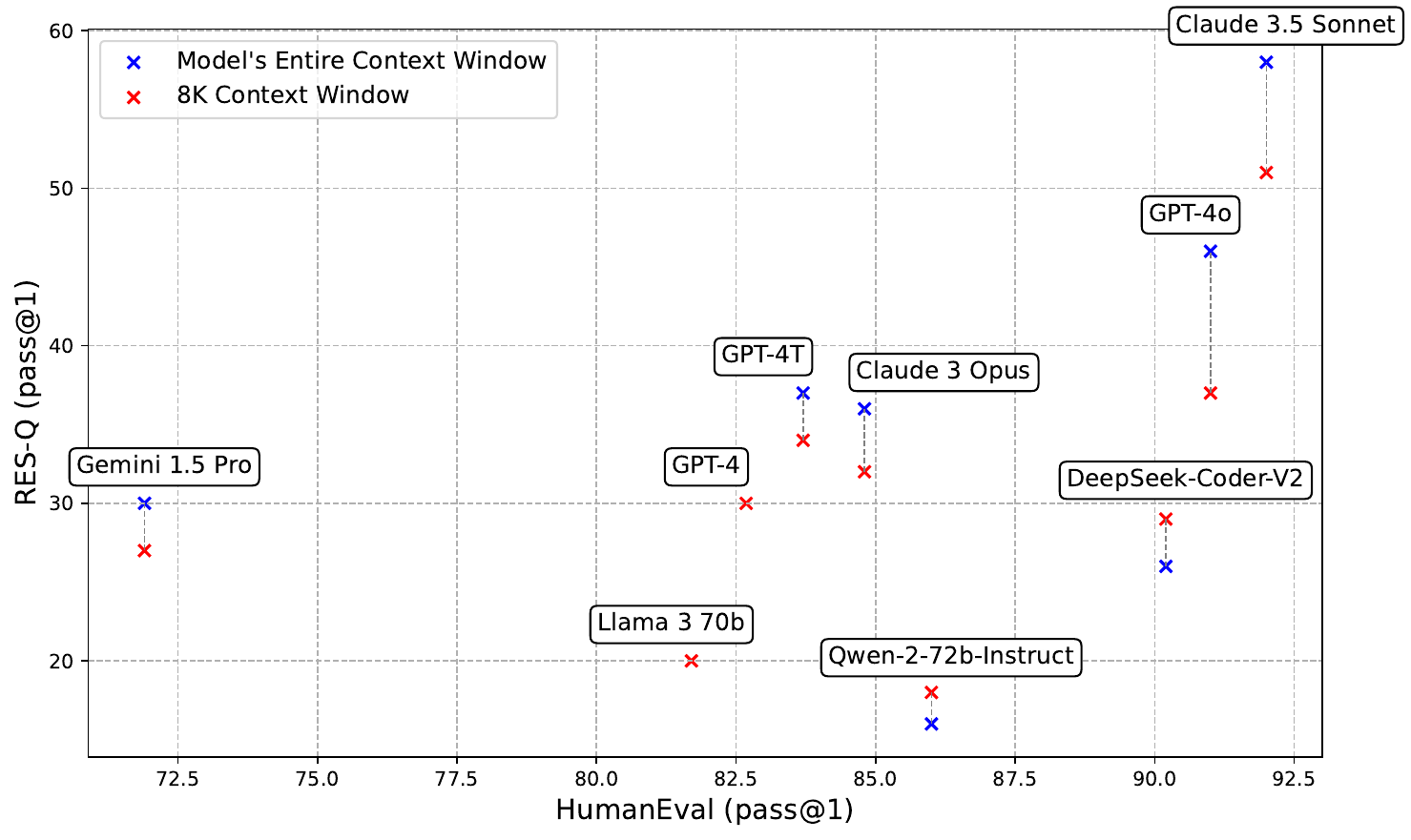}
        \caption{RES-Q vs HumanEval}
        \label{fig:humaneval-compare}
    \end{subfigure}%
    \hfill
    \begin{subfigure}[t]{0.48\textwidth}
        \centering
        \includegraphics[width=\linewidth]{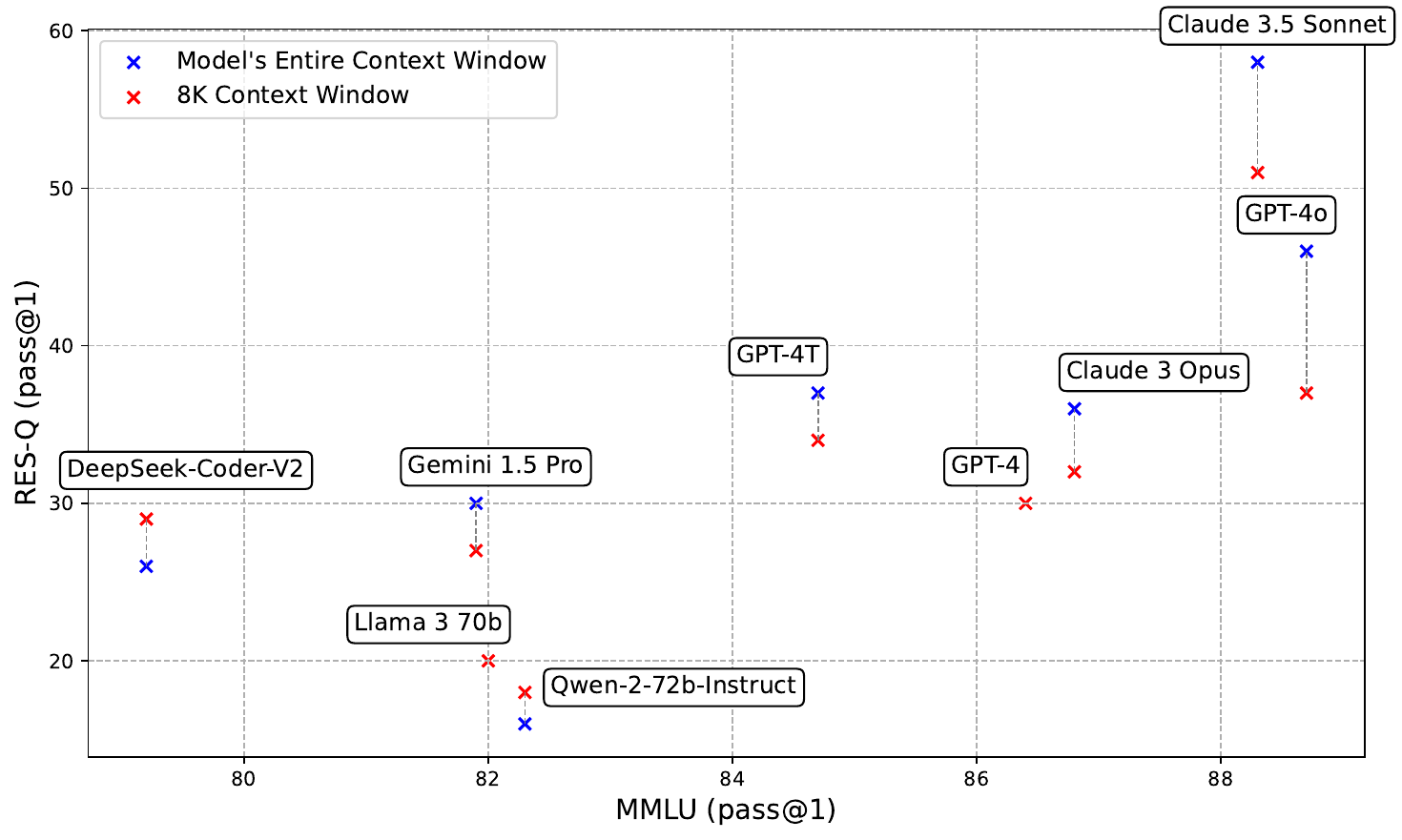}
        \caption{RES-Q vs MMLU}
        \label{fig:mmlu-compare}
    \end{subfigure}
    \caption{Model performances on \benchmark{} vs traditional LLM benchmarks. Note: The x-axes (HumanEval and MMLU scores) and y-axes (RES-Q scores) use different scales to better highlight differences between models.}
    \label{fig:benchmark-comparisons}
\end{figure}

In this section, we evaluate the performance of a set of state-of-the-art open and closed source LLMs on the \benchmark{} benchmark. 

\subsection{Experimental Setup}

To produce submissions for \benchmark{} tasks, we implemented an autonomous, multi-agent repository-editing system on top of Qurrent OS, our proprietary LLM agent development software. The system integrates code-editing tool use and an agent-based code retrieval system. Given an edit instruction and a Git repository, the system makes edits directly to the repository. We use the patch generated from the diff between the \textit{base commit} and the edits made by the system as our submission to the \benchmark{} submission environment.

We selected various state-of-the-art closed and open source LLMs for evaluation. We attempt all 100 \benchmark{} tasks once with each selected LLM acting in the repository-editing system and report their \textit{pass@1} scores.

Due to the wide range of context sizes in the models selected for evaluation (see \Cref{table:results}), we evaluated in two settings: (1) Unconstrained: each language agent is afforded its model's entire context window. (2) Constrained: each language agent is afforded 8K tokens of its model's context window. In the first setting, we evaluate the full capacity of each model and its ability to capitalize on its context window. In the second setting, we standardize window size, isolating evaluation of the model's per-token ability.




\subsection{Results}

\textbf{Constraining Context Window.} For the selected closed-source models, constraining the context window size to 8K tokens consistently decreased performance on \benchmark{}. The largest performance decreases by constraining context were observed in the two top-performing models: Claude 3.5 Sonnet (-9\%) and GPT-4o (-10\%). For the selected open-source models, constraining the context window consistently \textit{increased} performance. This suggests that these models may not be able to utilize their entire context window as effectively. Relative performance of models remained unchanged across the constrained and unconstrained settings.

\textbf{Closed vs Open Source LLMs.} With the exception of DeepSeek Coder V2 (29\% \textit{pass@1}) which outperformed Gemini Pro 1.5 (27\% \textit{pass@1}) in the constrained setting, closed-source LLMs outperformed open-source LLMs on RES-Q across the board. Most notably, Claude Sonnet 3.5 outperformed all other models including GPT-4o by 12\% \textit{pass@1} in the unconstrained setting and 15\% \textit{pass@1} in the constrained setting.

\textbf{Performance Correlation with Existing Benchmarks.} Our analysis identified two statistically significant performance correlations ($p < 0.05$) between RES-Q and the selected traditional benchmarks. When considering all selected LLMs, the correlation between RES-Q and MMLU was strong with a coefficient of 0.73 ($p = 0.00054$) while the correlation between RES-Q and HumanEval was not significant ($p = 0.054$). When considering only closed-source LLMs, the correlation between RES-Q and HumanEval was very strong with a coefficient of 0.86 ($p = 0.0016$) while the correlation between RES-Q and MMLU was not significant ($p = 0.065$). We did not identify significant performance correlations when considering only open-source models.

\textbf{Per-Task Token Usage.} We found that, despite its superior performance, Claude Sonnet 3.5 had the highest median token usage per \benchmark{} task in both the constrained and unconstrained settings. Qualitatively, this could indicate that Claude may make mistakes early on in a trajectory, but is able to recover and successfully realign to  a task. On the other hand, GPT-4o uses a disproportionately low amount of tokens per task ($\sim$ 50\% that of Sonnet), given its performance. This may indicate that it is better at making correct decisions early on in a trajectory.

\begin{figure}[h]
    \centering
    \includegraphics[width=0.88\linewidth]{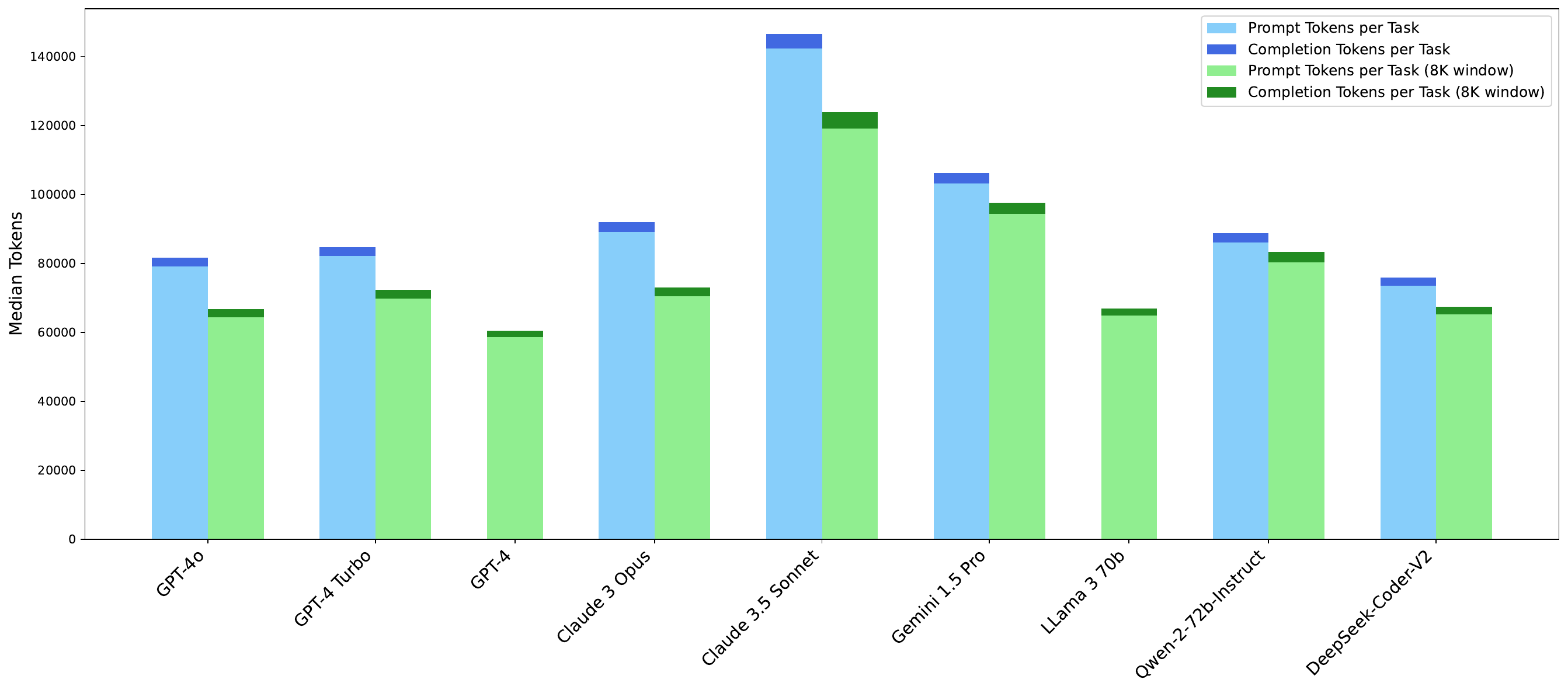}
    \caption{Median Input and Output Tokens Used Per Task by Model in each of the Constrained (8K window) and Unconstrained Settings.}
    \label{fig:example}
\end{figure}

\begin{table*}[t]
\centering
\renewcommand{\arraystretch}{1.1}
\begin{tabular}{|c|c|c|c|c|c|}
\hline
\multicolumn{2}{|c|}{Model} & \multicolumn{4}{c|}{Benchmark Results (pass@1)} \\
\hline
Name & Context Window & HumanEval & MMLU & \benchmark{} & \benchmark{} (8K)  \\
\hline
\multicolumn{6}{|c|}{Closed Models} \\
\hline
Claude 3.5 Sonnet  & 200K & 92.0 & 88.3 & \textbf{58} & \textbf{51} \\
GPT-4o  & 128K & 91 & 88.7 & 46 & 36\\
GPT-4T  & 128K & 88.2 & 86.5 & 37 & 34 \\
Claude 3 Opus  & 200K & 84.8 & 86.8 & 36 & 32 \\
GPT-4   & 8K & 82.68 & 86.4 & 30 & 30  \\
Gemini 1.5 Pro & 1M & 71.9 & 81.9 & 30 & 27 \\
\hline
\multicolumn{6}{|c|}{Open Models} \\
\hline
DeepSeek Coder V2 Instruct  & 128K & 90.2 & 79.2 & \textbf{26} & \textbf{29} \\
Llama 3 70b  & 8K & 81.7 & 82 & 20 & 20 \\
Qwen2-72B Instruct  & 32K & 86.0 & 82.3 & 16 & 18 \\
\hline
\end{tabular}
\caption{
  Performance of selected LLMs on \benchmark{} vs traditional NLP benchmarks along with their context window size. Results reported for \benchmark{} are the product of a language agent system built on Qurrent OS.
  For details on the API endpoints of closed models, see Appendix~\ref{appx:api-endpoints}. 
}
\label{table:results}
\end{table*}



\bibliography{iclr2024_conference}
\bibliographystyle{iclr2024_conference}

\appendix
\section{Appendix}\label{sec:appx}
\subsection{Example RES-Q Task}\label{appx:example-task}
\begin{tcolorbox}[colback=purple!10!white, colframe=purple!50!black, title=Instruction]
Enhance the HTTPClient class in http.py to allow editing various application details via Discord's API. Implement the \texttt{edit\_application\_info} method to send a PATCH request to the /applications/@me endpoint. This method should accept \texttt{reason} and \texttt{payload}, filter the payload for specific valid keys like 'custom\_install\_url', 'description', etc., and use \texttt{self.request} to send the PATCH request with the filtered payload and reason. Ensure the method returns the response from the API call.
\end{tcolorbox}

\begin{tcolorbox}[colback=gray!10!white, colframe=gray!50!black, title=Ground Truth Patch]
\begin{lstlisting}
diff --git a/discord/http.py b/discord/http.py
--- a/discord/http.py
+++ b/discord/http.py
@@ -2456,6 +2456,22 @@ class HTTPClient:
     def application_info(self) -> Response[appinfo.AppInfo]:
         return self.request(Route('GET', '/oauth2/applications/@me'))
 
+    def edit_application_info(self, *, reason: Optional[str], payload: Any) -> Response[appinfo.AppInfo]:
+        valid_keys = (
+            'custom_install_url',
+            'description',
+            'role_connections_verification_url',
+            'install_params',
+            'flags',
+            'icon',
+            'cover_image',
+            'interactions_endpoint_url',
+            'tags',
+        )
+
+        payload = {k: v for k, v in payload.items() if k in valid_keys}
+        return self.request(Route('PATCH', '/applications/@me'), json=payload, reason=reason)
+

\end{lstlisting}
\end{tcolorbox}

\subsection{API Endpoints for Closed Models}\label{appx:api-endpoints}

\begin{table*}[h]
\centering
\renewcommand{\arraystretch}{1.1}
\begin{tabular}{|c|c|}
\hline
Model & API Endpoint \\
\hline
GPT-4o & gpt-4o \\
GPT-4T & gpt-4-turbo-2024-04-09 \\
GPT-4 & gpt-4-0613 \\
Claude 3 Opus & claude-3-opus-20240229 \\
Claude 3.5 Sonnet & claude-3-5-sonnet-20240620\\
Gemini 1.5 Pro & gemini-1.5-pro-001 \\
\hline
\end{tabular}
\caption{
  API endpoints for the closed models used in our evaluations.
}
\label{table:api-endpoints}
\end{table*}

\subsection{Submission Environment}\label{appx:submission-env}
\includegraphics[width=\linewidth]{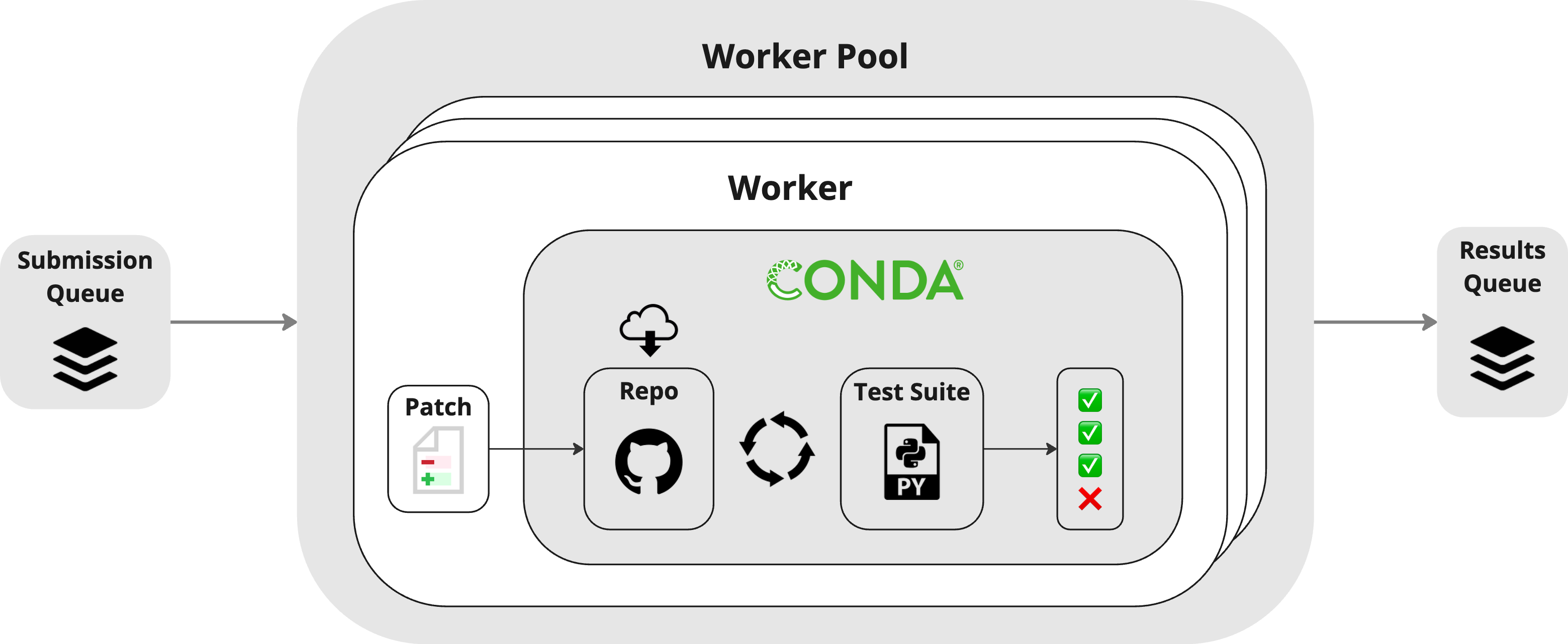}

\end{document}

%% file: math_commands.tex
\usepackage{amsmath,amsfonts,bm}

\def\eqref#1{equation~\ref{#1}}

\def\1{\bm{1}}

\DeclareMathAlphabet{\mathsfit}{\encodingdefault}{\sfdefault}{m}{sl}
\SetMathAlphabet{\mathsfit}{bold}{\encodingdefault}{\sfdefault}{bx}{n}













%% file: iclr2024_conference.bbl
\begin{thebibliography}{13}
\providecommand{\natexlab}[1]{#1}
\providecommand{\url}[1]{\texttt{#1}}
\expandafter\ifx\csname urlstyle\endcsname\relax
  \providecommand{\doi}[1]{doi: #1}\else
  \providecommand{\doi}{doi: \begingroup \urlstyle{rm}\Url}\fi

\bibitem[Cassano et~al.(2024)Cassano, Li, Sethi, Shinn, Brennan-Jones, Ginesin, Berman, Chakhnashvili, Lozhkov, Anderson, and Guha]{cassano2024edit}
Federico Cassano, Luisa Li, Akul Sethi, Noah Shinn, Abby Brennan-Jones, Jacob Ginesin, Edward Berman, George Chakhnashvili, Anton Lozhkov, Carolyn~Jane Anderson, and Arjun Guha.
\newblock Can it edit? evaluating the ability of large language models to follow code editing instructions, 2024.

\bibitem[Chen et~al.(2024)Chen, Lin, Zeng, Zan, Wang, Cheshkov, Sun, Yu, Dong, Aliev, Wang, Cheng, Liang, Ma, Bian, Xie, and Wang]{chen2024coder}
Dong Chen, Shaoxin Lin, Muhan Zeng, Daoguang Zan, Jian-Gang Wang, Anton Cheshkov, Jun Sun, Hao Yu, Guoliang Dong, Artem Aliev, Jie Wang, Xiao Cheng, Guangtai Liang, Yuchi Ma, Pan Bian, Tao Xie, and Qianxiang Wang.
\newblock Coder: Issue resolving with multi-agent and task graphs, 2024.

\bibitem[Gauthier(2023)]{aider}
Paul Gauthier.
\newblock Aider, 2023.
\newblock URL \url{https://aider.chat/}.

\bibitem[Hendrycks et~al.(2021)Hendrycks, Burns, Basart, Zou, Mazeika, Song, and Steinhardt]{hendrycks2021measuring}
Dan Hendrycks, Collin Burns, Steven Basart, Andy Zou, Mantas Mazeika, Dawn Song, and Jacob Steinhardt.
\newblock Measuring massive multitask language understanding, 2021.

\bibitem[Hong et~al.(2023)Hong, Zhuge, Chen, Zheng, Cheng, Zhang, Wang, Wang, Yau, Lin, Zhou, Ran, Xiao, Wu, and Schmidhuber]{hong2023metagpt}
Sirui Hong, Mingchen Zhuge, Jonathan Chen, Xiawu Zheng, Yuheng Cheng, Ceyao Zhang, Jinlin Wang, Zili Wang, Steven Ka~Shing Yau, Zijuan Lin, Liyang Zhou, Chenyu Ran, Lingfeng Xiao, Chenglin Wu, and Jürgen Schmidhuber.
\newblock Metagpt: Meta programming for a multi-agent collaborative framework, 2023.

\bibitem[Jimenez et~al.(2024)Jimenez, Yang, Wettig, Yao, Pei, Press, and Narasimhan]{jimenez2024swebench}
Carlos~E. Jimenez, John Yang, Alexander Wettig, Shunyu Yao, Kexin Pei, Ofir Press, and Karthik Narasimhan.
\newblock Swe-bench: Can language models resolve real-world github issues?, 2024.

\bibitem[Osika(2023)]{gpt-engineer}
Anton Osika.
\newblock gpt-engineer, 2023.
\newblock URL \url{https://gpt-engineer.readthedocs.io}.

\bibitem[Ott et~al.(2022)Ott, Barbosa-Silva, Blagec, Brauner, and Samwald]{Ott2022MappingGD}
Simon Ott, Adriano Barbosa-Silva, Kathrin Blagec, Janina Brauner, and Matthias Samwald.
\newblock Mapping global dynamics of benchmark creation and saturation in artificial intelligence.
\newblock \emph{Nature Communications}, 13, 2022.
\newblock URL \url{https://api.semanticscholar.org/CorpusID:247318891}.

\bibitem[Sainz et~al.(2023)Sainz, Campos, Garc{\'\i}a-Ferrero, Etxaniz, de~Lacalle, and Agirre]{sainz-etal-2023-nlp}
Oscar Sainz, Jon Campos, Iker Garc{\'\i}a-Ferrero, Julen Etxaniz, Oier~Lopez de~Lacalle, and Eneko Agirre.
\newblock {NLP} evaluation in trouble: On the need to measure {LLM} data contamination for each benchmark.
\newblock In Houda Bouamor, Juan Pino, and Kalika Bali (eds.), \emph{Findings of the Association for Computational Linguistics: EMNLP 2023}, pp.\  10776--10787, Singapore, December 2023. Association for Computational Linguistics.
\newblock \doi{10.18653/v1/2023.findings-emnlp.722}.
\newblock URL \url{https://aclanthology.org/2023.findings-emnlp.722}.

\bibitem[{Significant Gravitas}(2024)]{AutoGPT}
{Significant Gravitas}.
\newblock Autogpt, 2024.
\newblock URL \url{https://agpt.co}.

\bibitem[Suzgun et~al.(2022)Suzgun, Scales, Schärli, Gehrmann, Tay, Chung, Chowdhery, Le, Chi, Zhou, and Wei]{suzgun2022challenging}
Mirac Suzgun, Nathan Scales, Nathanael Schärli, Sebastian Gehrmann, Yi~Tay, Hyung~Won Chung, Aakanksha Chowdhery, Quoc~V. Le, Ed~H. Chi, Denny Zhou, and Jason Wei.
\newblock Challenging big-bench tasks and whether chain-of-thought can solve them, 2022.

\bibitem[Weng(2023)]{weng2023agent}
Lilian Weng.
\newblock Llm-powered autonomous agents.
\newblock \emph{lilianweng.github.io}, Jun 2023.
\newblock URL \url{https://lilianweng.github.io/posts/2023-06-23-agent/}.

\bibitem[Xi et~al.(2023)Xi, Chen, Guo, He, Ding, Hong, Zhang, Wang, Jin, Zhou, Zheng, Fan, Wang, Xiong, Zhou, Wang, Jiang, Zou, Liu, Yin, Dou, Weng, Cheng, Zhang, Qin, Zheng, Qiu, Huang, and Gui]{xi2023rise}
Zhiheng Xi, Wenxiang Chen, Xin Guo, Wei He, Yiwen Ding, Boyang Hong, Ming Zhang, Junzhe Wang, Senjie Jin, Enyu Zhou, Rui Zheng, Xiaoran Fan, Xiao Wang, Limao Xiong, Yuhao Zhou, Weiran Wang, Changhao Jiang, Yicheng Zou, Xiangyang Liu, Zhangyue Yin, Shihan Dou, Rongxiang Weng, Wensen Cheng, Qi~Zhang, Wenjuan Qin, Yongyan Zheng, Xipeng Qiu, Xuanjing Huang, and Tao Gui.
\newblock The rise and potential of large language model based agents: A survey, 2023.

\end{thebibliography}
